\documentclass[10pt,twocolumn,letterpaper]{article}
\usepackage{authblk}
\usepackage{xcolor}
\usepackage{multicol,multirow}
\usepackage{overpic}
\usepackage{float}
\usepackage{wrapfig}
\usepackage{amsmath}
\usepackage{graphicx}
\usepackage{amssymb}
\usepackage{booktabs}
\usepackage{afterpage}

\newcommand{\setfootnotemark}{%
  \refstepcounter{footnote}%
  \footnotemark[\value{footnote}]}

\newcommand{\pc}[0]{\mathbf{P}}

\newcommand{\classes}[0]{C}
\newcommand{\objclass}[0]{c}
\newcommand{\objclassgt}[0]{\widehat{C}_{P}}

\newcommand{\objtemplate}[0]{\mathbf{T}_{\objclass}}

\newcommand{\objkeyp}[0]{\mathbf{K}_{\objclass}}

\newcommand{\objcenter}[0]{\mathbf{o}_{P}}
\newcommand{\objcentergt}[0]{\widehat{\mathbf{o}}_{P}}
\newcommand{\pclocal}[0]{\pc_L}
\newcommand{\featobjcenter}[0]{\mathbf{F}_{o}}
\newcommand{\featpclocal}[0]{\mathbf{F}_{P_L}}

\newcommand{\offset}[0]{\mathbf{S}_{K}}
\newcommand{\offsetgt}[0]{\widehat{\mathbf{S}}_{{K}}}

\newcommand{\ourout}[0]{\mathbf{T}'}

\DeclareMathOperator*{\argminB}{argmin} 

\usepackage[pagenumbers]{cvpr} %

\usepackage[pagebackref,breaklinks,colorlinks]{hyperref}
\usepackage[capitalize]{cleveref}
\usepackage[accsupp]{axessibility} 
\crefname{section}{Sec.}{Secs.}
\Crefname{section}{Section}{Sections}
\Crefname{table}{Table}{Tables}
\crefname{table}{Tab.}{Tabs.}

\def\confName{CVPR}
\def\confYear{2023}

\begin{document}
\title{Object pop-up: Can we infer 3D objects and their poses from human interactions alone?}

\makeatletter
\renewcommand\AB@affilsepx{ , \protect\Affilfont}
\makeatother

\author[1,2]{Ilya A. Petrov}
\author[1,2]{Riccardo Marin}
\author[1,3]{Julian Chibane}
\author[1,2,3]{Gerard Pons-Moll}
\affil[1]{\small University of T\"ubingen, Germany}

\makeatletter
\renewcommand\AB@affilsepx{\protect\\\protect\Affilfont}
\makeatother

\affil[2]{\small T\"ubingen AI Center, Germany}
\affil[3]{\small Max Planck Institute for Informatics, Saarland Informatics Campus, Germany}
\affil[ ]{\tt\small \{i.petrov, gerard.pons-moll\}@uni-tuebingen.de, riccardo.marin@mnf.uni-tuebingen.de, jchibane@mpi-inf.mpg.de}

\twocolumn[{%
\renewcommand\twocolumn[1][]{#1}%
\maketitle
\begin{center}
    \centering
    \captionsetup{type=figure}
    \vspace{2pt}

    \begin{overpic}
    [trim=0cm 0cm 0cm 0cm,clip,width=1.0\textwidth]{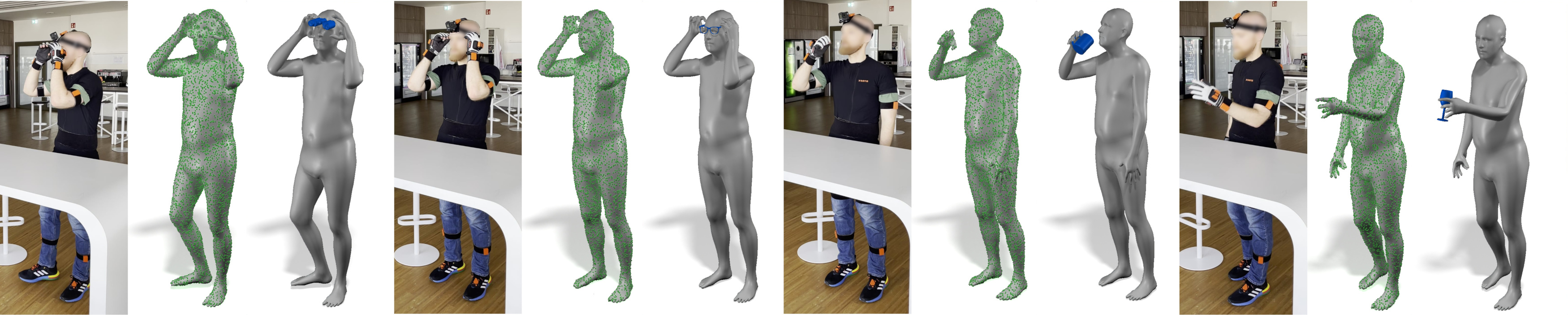}
    \scriptsize
    \put(0.7,22){Ref. Image}
    \put(1.2,20.5){(not used)}
    
    \put(11,22){Input}
    \put(9,20.5){Point Cloud}
    
    \put(18.5,21){Result}
        
    \put(26,22){Ref. Image}
    \put(26.5,20.5){(not used)}
    
    \put(36.5,22){Input}
    \put(34.5,20.5){Point Cloud}
    
    \put(44,21){Result}    
  
    \put(51,22){Ref. Image}
    \put(51.5,20.5){(not used)}
    
    \put(61,22){Input}
    \put(59.5,20.5){Point Cloud}
    
    \put(69,21){Result}        

    \put(76,22){Ref. Image}
    \put(76.5,20.5){(not used)}
    
    \put(86,22){Input}
    \put(84.5,20.5){Point Cloud}
    
    \put(93,21){Result}     
    
    \end{overpic}

    \caption{\textbf{Human-Centric prediction of Human-Object Interaction.} We explore Human-Object interaction when only the cues of the human pose interacting with an unobserved object are available ('Input'). We propose a neural model that, for the first time, can infer the location of the object ('Result') from such input. This is possible even when our subject simulates the interactions with the object ('Reference Image').}
    \label{fig:results_imu}
    \vspace{3pt}
\end{center}%
}]

\begin{abstract}
The intimate entanglement between objects affordances and human poses is of large interest, among others, for behavioural sciences, cognitive psychology, and Computer Vision communities. 
In recent years, the latter has developed several object-centric approaches: starting from items, learning pipelines synthesizing human poses and dynamics in a realistic way, satisfying both geometrical and functional expectations. 
However, the inverse perspective is significantly less explored: Can we infer 3D objects and their poses from human interactions alone? 
Our investigation follows this direction, showing that a generic 3D human point cloud is enough to pop up an unobserved object, even when the user is just imitating a functionality (e.g., looking through a binocular) without involving a tangible counterpart. 
We validate our method qualitatively and quantitatively, with synthetic data and sequences acquired for the task, showing applicability for XR/VR. The code is available at: \href{https://github.com/ptrvilya/object-popup}{https://github.com/ptrvilya/object-popup}.

\end{abstract}
\section{Introduction}
\label{sec:introduction}
\begin{figure}[!t]
    \centering
    \captionsetup{type=figure}
    \begin{overpic}
    [trim=0cm 0cm 0cm 0cm,clip,width=1\linewidth]{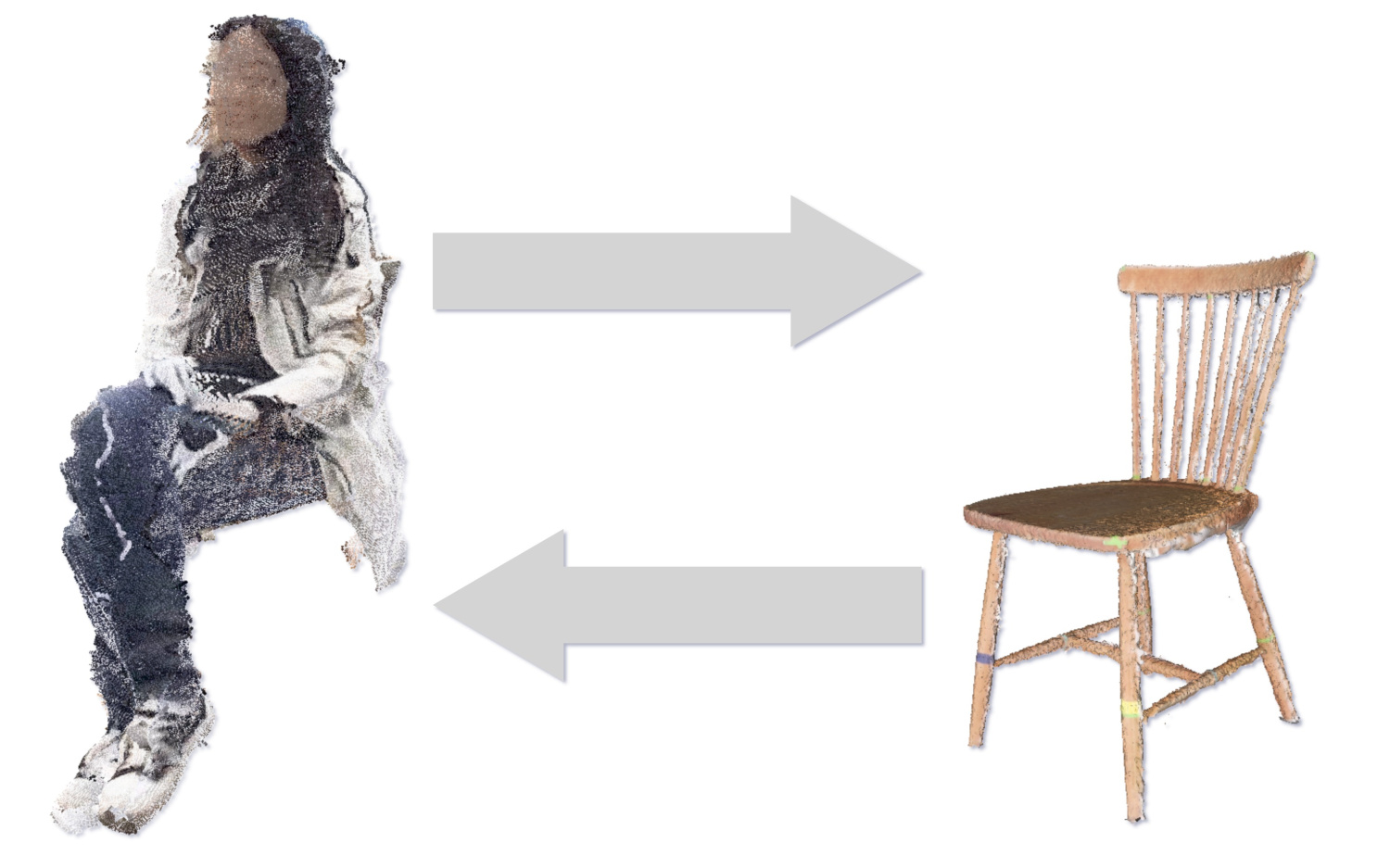}
    \put(43,48){\textbf{Our}}
    \put(36,41.5){Human-Centric}
    \put(36,17.5){Object-Centric}
    \put(36,10){Previous Work}
    \end{overpic}
    \caption{\textbf{Human-Centric \vs Object-Centric inference.} A common perspective in the prior work is to infer affordances or human pose starting from an object. We explore the inverse, Human-Centric perspective of the human-object interaction relationship: given the human, we are interested in predicting the object.}
    \label{fig:human_centric}
    \vspace{-10pt}
\end{figure}

Complex interactions with the world are among the unique skills distinguishing humans from other living beings. 
Even though our perception might be imperfect (we cannot hear ultrasonic sounds or see ultraviolet light \cite{SCANES20181}), our cognitive representation is enriched with a functional perspective, i.e., potential ways of interacting with objects or, as introduced by Gibson and colleagues, the affordance of the objects \cite{gibson1982concept}. 
Several behavioural studies confirmed the centrality of this concept \cite{carlson1999effects, booth2002object, oakes2008function}, which plays a fundamental role also for kids' development \cite{booth2002object}. 
Computer Vision is well-aware that the function of an object complements its appearance \cite{grabner2011what}, and exploited this in tasks like human and object reconstruction \cite{yao2013discovering}. 
Previous literature approaches the interaction analysis from an object perspective (i.e., given an object, analyze the human interaction)\cite{grady2021contactopt, cao2021reconstructing, ye2022what}, building object-centric priors \cite{zhou2022toch}, generating realistic grasps given the object \cite{karunratanakul2020grasping, christen2022dgrasp}, reconstructing hand-object interactions \cite{grady2021contactopt, cao2021reconstructing, ye2022what}.
Namely, objects induce functionality, so a human interaction (e.g., a mug suggests a drinking action; an handle a grasping one). 
For the first time, our work reverts the perspective, suggesting that analyzing human motion and behaviour is naturally a human-centric problem (i.e., given a human interaction, what kind of functionality is it suggesting, \cref{fig:human_centric}). 
Moving the first step in this new research direction, we pose a fundamental question: \emph{Can we infer 3D objects and their poses from human interactions alone?}

\begin{figure}
\vspace{0.1cm}

    \begin{overpic}
    [trim=0cm 0cm 0cm 0cm,clip,width=0.95\linewidth]{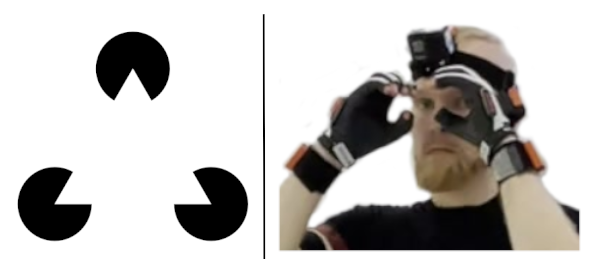}
    \put(74,-1.5){B)}
    \put(22,-1.5){A)}
    \end{overpic}
    \vspace{0.05cm}
 
\caption{\label{fig:hand} \textbf{Reification principle.} Gestalt theory suggests our perception considers the whole before the single parts (A). This served as inspiration for our work: the object can arise considering the body parts as a whole (B).}
\end{figure}

At first sight, the problem seems particularly hard and significantly under-constrained since several geometries might fit the same action. 
However, the human body complements this information in several ways: physical relations, characteristic poses, or body dynamics serve as valuable proxies for the involved functionality, as suggested by \cref{fig:results_imu}. 
Such hints are so powerful that we can easily imagine the kind of object and its location, even if such an object does not exist. 
Furthermore, even if the given pose might fit several possible solutions, our mind naturally comes to the most natural one indicated by the observed \emph{behaviour}. 
It also suggests that solely focusing on the contact region (an approach often preferred by previous works) is insufficient in this new viewpoint. The reification principle of Gestalt psychology \cite{lehar2003world} highlights that "the whole" arises from "the parts" and their relation. Similarly, in~\cref{fig:hand}, the hand grasps in B) pop up a binocular in our mind because we naturally consider the relationship with the other body parts.

Finally, moving to a human-centric perspective in human-object interaction is critical for human studies and daily-life applications. 
Modern systems for AR/VR \cite{guzov23ireplica}, and digital interaction \cite{harih2013toolhandle} are both centered on humans, often manipulating objects that do not have a real-world counterpart. Learning to decode an object from human behaviour enables unprecedented applications.

To answer our question, we deploy a first straightforward and effective pipeline to "pop up" a rigid object from a 3D human point cloud. 
Starting from the input human point cloud and a class, we train an end-to-end pipeline to infer object location. 
In case a temporal sequence of point clouds is available, we suggest post-processing to avoid jittering and inconsistencies in the predictions, showing the relevance of this information to handle ambiguous poses.
We show promising results on previously unaddressed tasks in digital and real-world scenes.
Finally, our method allows us to analyze different features of human behaviour, highlight their contribution to object retrieval, and point to exciting directions for future works. 

In summary, our main contributions are:
\begin{enumerate}
    \item We formulate a novel problem, changing the perspective taken by previous works in the field, and open to a yet unexplored research direction;
    \item We introduce a method capable of predicting the object starting from an input human point cloud;
    \item We analyze different components of the human-object relationship: the contribution of different pieces of interactions (hands, body, time sequence), the point-wise saliency of input points, and the confusion produced by objects with similar functions.
\end{enumerate}

\section{Related Work}
\label{sec:related_work}

\subsection{Object Functionality}
In the core of human perception of objects, functionality complements physical appearance, enhancing our perception. 
Gibson \cite{gibson1982concept} introduced the idea that humans use \emph{affordances} of objects for perception. 
Affordance can be defined as: "an intrinsic property of an object, allowing an action to be performed with the object" \cite{kjellstrom2011visual}.

From a Computer Vision perspective, object functionality supports several tasks such as scene analysis \cite{grabner2011what}, object classification \cite{kjellstrom2011visual, deng2021affordancenet}, object properties inferring \cite{zheng2021inferring}, and it is also possible to learn object-specific human interaction models from 2D images\cite{yao2013discovering}.
These works suggest an intimate entanglement between the action and the object itself.

\subsection{Human-Object interaction}
Modelling environment-aware humans and their interactions in 3D is one of the most recent challenges to creating virtual humans. 
We see two main lines of work there: hand-focused and full-body ones.

\paragraph{Hand-Object Interaction.} Several works tackle the problem of human interaction, focusing solely on the hands \cite{chao2021dexycb}. 
This has been done starting from 2D \cite{corona2020ganhand, ehsani2020use, grady2021contactopt, hasson2019learning, yang2021cpf, cao2021reconstructing, karunratanakul2020grasping, hasson2021towards, hampali2022keypoint, ye2022what}, 2.5D \cite{brahmbhatt2019contactdb, brahmbhatt2020contactpose}, and 3D data \cite{brahmbhatt2019contactgrasp, karunratanakul2020grasping, taheri2020grab}. 
Particularly promising seems the application of well-designed priors for the motion \cite{zhou2022toch}. 
The class of objects involved in these works are mainly limited to graspable ones. 
We argue that interactions involving different body parts are common in everyday life, more attractive from an applicative perspective, and more challenging. 
Moreover, full-body context is crucial in reconstructing even grasp interactions since body pose contains information on an object's properties, e.g. pounding with a hammer affects the whole posture to support the action.

\paragraph{Fully-Body Interaction.} On this line, several works focus on the interaction between a human and a scene  \cite{yi2022humanaware, li2022mocapdeform, chen2019holisticpp, hassan2019resolving, weng2021holistic, zanfir2018monocular, zhang2021learning,huang2022rich}. 
Also, in this case, priors can be used to regularize the motion \cite{savva2016pigraphs}.
Several datasets are also available to study the interaction between a human and a single object. For example, recent BEHAVE \cite{bhatnagar2022behave}, GRAB \cite{taheri2020grab}, and InterCap \cite{huang2022intercap} capture full-body interactions with diverse objects. Works address the task of humans interactions reconstruction from different kinds of data sources like single image \cite{xie2022chore,zhang2020perceiving}, video \cite{su2021robustfusion, dabral2021gravityawarea, xie2023vistracker}, and multi-view capturing \cite{bhatnagar2022behave, jiang2022neuralhofusion, sun2021neural}, and synthetization of them as well \cite{zhang2022couch, cao2020long, hassan2021stochastic,wu2022saga,taheri2022goal,wan2022learn, mandikal2021learning,zhang2019predicting}.  
However, in all these works, we observe a general object-centric perspective: given a scene or an object, they aim to recreate the humans interacting with them. We argue that the significantly less explored complementary one has more concrete applications in daily life, especially in VR/XR contexts where the human is central to the system.

\begin{figure*}[!ht]
\centering
    \begin{overpic}[trim=0cm 0cm 0cm 0cm,clip,width=0.92\linewidth]{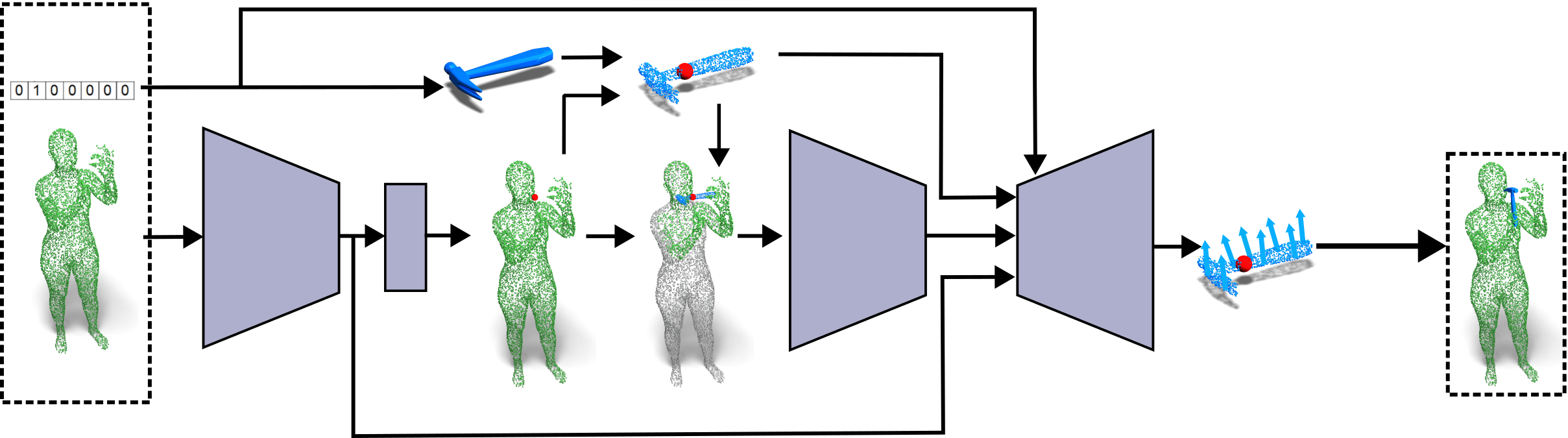}
    \scriptsize
    \put(1,26){Object Class}
    \put(4,24){$\objclass$}
    
    \put(1,4){Point Cloud}
    \put(4,2.5){$\mathbf{P}$}    

    \put(28,26){Template}
    \put(28,24.4){$\objtemplate$}    
    
    \put(40,26){Key points}
    \put(40,24.4){$\objkeyp$}       
    
    \put(30,2.2){Object Center}
    \put(33,0.7){$\objcenter$}        

    \put(41,2.2){Local Neigh.}
    \put(44,0.7){$\pclocal$}  
    
    \put(77,6){Key points}
    \put(78.3,4){offsets}    
    \put(78.8,2){$\offset$}  
    
    \put(85,16){Template}  
    \put(85.5,14){Fitting}  
    
    \put(94,1){Output}  
    
    \put(13.5,12){PointNet++}  
    \put(51.5,12){PointNet++}  
    \put(67,12){Decoder} 
    
    \put(60.5,16){$\objkeyp$}
    \put(60.4,8.5){$\featobjcenter$} 
    \put(59.8,13.7){$\featpclocal$} 
    \put(66.2,19){$\objclass$} 
    
    \put(25.6,14){\rotatebox{-90}{MLP}}
    \end{overpic}
    
\caption{\label{fig:pipeline} \textbf{Object pop-up.} Our method predicts the position of an object, starting only from an input point cloud and an object class. It relies on a careful problem decomposition in several sub-tasks, extracting features that involve the entire human body and relations between body parts near the object.}
    \vspace{-15pt}
\end{figure*}

\subsection{Human-centric perspective}
While the general trend mainly focuses on the surrounding environment and objects, there is a growing interest and availability of egocentric tools for humans \cite{guzov2021human, zhang2021egobody} also interacting with objects \cite{kwon2021h2o, guzov23ireplica,liu2022hoi4d}. They provide a subjective view and a valuable paradigm for several applications, like letting a user interact with objects in the digital world. 
Recent works also involve more sophisticated devices \cite{sundaram2019learning, sun2022augmented}, while they are still far from applicability. 
In a similar direction to our work, others propose to recover objects arrangements in a room starting from the human motion \cite{nie2022pose2room}, to hallucinate a coherent 2D image from a human pose \cite{brooks2022hallucinating}, or predicting physical properties of the objects (e.g., the weight of a box) from human joints \cite{zheng2021inferring}. While the principle inspires us, our study significantly differs: we focus on object pose and its spatial relation with humans, starting solely from unordered point clouds.

\section{Method}
\label{sec:method}
This section describes our setting and the main components of our methodology, both at inference and training time.
An overview of our pipeline can be found in \cref{fig:pipeline}.

\subsection{Object Pop-Up}

\paragraph{Input.} Our method starts from a single human point cloud $\pc \in \mathbb{R}^{N_P \times 3}$ with $N_P$ points and a hot-encoded object class $\objclass$. The input point cloud can result from 3D/4D scans, IMUs template fitting, or any other shape-from-X approach. Regardless of the point cloud source, we remark that no further information is used apart from the 3D coordinates of the points. We represent each object as $1500$ key points $\objkeyp$ uniformly sampled on the template mesh.

\paragraph{Object Center.} Training a model to predict an object pose from a human point cloud poses several challenges. Such a task requires the network to understand the location of different body parts and their subtle relations while jointly developing a sense of its spatial relationship with the human. Empirically, we observed that this is only feasible by carefully deconstructing the problem and designing different features to ease the learning process.
As the first step to decompose this problem, we train a PointNet++ architecture \cite{qi2017pointnet, qi2017pointnetpp} to predict the object center $\objcenter$ starting from $\pc$. At training time, this is supervised with an L2 loss against the ground truth center $\objcentergt$:
\begin{equation}
    L_{\objcenter}(\pc) = \| \objcenter - \objcentergt \|_2^2.
\end{equation}
Solving this task provides a good initialization for the object pose, and we move the key points associated with the input class $\objclass$ to the predicted center  $\objcenter$.
Also,  the features $\featobjcenter \in \mathbb{R}^{512}$ extracted by the network encode import information on the whole human body.

\paragraph{Local Neighbourhood.} The center prediction module can be further exploited using the nearest human regions. Intuitively, considering the closest body parts is essential to infer a contact relationship, but also the influence of body parts not directly in touch with the object (e.g., head orientation while using binoculars). 
To learn these connections, we consider the centered key points together with the $3000$ closest points of the input human point cloud: $\pclocal = KNN(\pc, \objcenter)$. We pass these two sets as a unique point cloud to a PointNet++ network, obtaining a new set of per-point features $\featpclocal \in \mathbb{R}^{128}$.

\paragraph{Object displacement.} To predict the object's final position, we empirically observed that directly predicting a rotation and a translation is not a good solution. Inspired by recent works that suggest a point-wise offset prediction to recover 3D human shapes \cite{corona2022learned}, we apply a similar approach to our task.  Our goal is to predict a point-wise shift $\offset$ for the $\objkeyp$ vertices to align them to the target pose.
We append the features $\featobjcenter$, $\featpclocal$, the one-hot encoding of the object class $\objclass$, and a positional encoding to the centered key points, and we pass them to a decoder.
At training time, we consider the following loss:
\begin{equation}
    L_{off}(\objkeyp, \featobjcenter, \featpclocal, \objclass) = \|\offset - \offsetgt \|_F^2.
    \label{eq:loss_off}
\end{equation}

The network is then trained end-to-end using:
\begin{equation}
    L = L_{\objcenter} + \alpha L_{off}.
\end{equation}
The weighting coefficient is $\alpha = 10$.

\subsection{Template fitting}
\paragraph{Procrustes alignment.} The point-wise offset produced by the network potentially distorts the key points structure in a non-rigid way. To recover the desired global rigid transformation, we rely on a Procrustes alignment \cite{gower1975generalized}. This procedure takes as input two point clouds and returns the rotation $\mathbf{R}$ and the translation $\mathbf{t}$ to minimize the L2 distances of the points:
\begin{equation}
    P(\cdot, \cdot) \rightarrow (\mathbf{R}, \mathbf{t}).
\end{equation}

We apply this to the template key points and their configuration obtained with our network:
\begin{equation}
    P(\objkeyp, \objkeyp + \objcenter + \offset) \rightarrow (\mathbf{R}, \mathbf{t}).
\end{equation}

Finally, we recover the desired object pose as:
\begin{equation}
    \ourout = \mathbf{R} \objtemplate + \mathbf{t}
\end{equation}

\paragraph{Time Smoothing.} While our pipeline is designed to work with a single point cloud as input, considering the temporal evolution of interaction is often crucial, shaping the context of the individual poses. If a temporal sequence of point clouds is available, we provide a post-processing smoothing technique to take advantage of this further information. 
After running our method for each frame, we smooth the centering prediction across the sequence using a Gaussian kernel. Later, we will discuss a variation of our approach that also predicts the object class. 
In that case, we consider the most frequent class prediction over the whole set of frames to fix a class for the sequence.

\section{Experiments}
\label{sec:experiments}
In this Section, we will describe the datasets used for training and testing our method. 
Then, we will present the baseline and the evaluation metrics. 
Finally, we will provide validation of our method as well as an analysis of its extended version that allows object class prediction.

\begin{table*}[ht!]
  \small
  \centering
  \begin{tabular}{lccccccccc}
    \multirow{2}{*}{\bf Methods} & \multicolumn{2}{c}{\bf GRAB} & \multicolumn{2}{c}{\bf BEHAVE} &  \multicolumn{2}{c}{\bf BEHAVE-Raw} \\
    \cline{2-7}
                                & $E_c\downarrow$  & $E_{v2v}\downarrow$ & $E_c\downarrow$ & $E_{v2v}\downarrow$ & $E_c\downarrow$ & $E_{v2v}\downarrow$ \\
    \hline
    NN               & 0.0362           & 0.1445           & 0.0802  & 0.3445 & - & -\\
    \textbf{Ours}    & \textbf{0.0237}  & \textbf{0.0943} & \textbf{0.0663} & \textbf{0.2900} & \textbf{0.0806}  & \textbf{0.3143} \\
  \end{tabular}
  \caption{ \textbf{Comparison with the baseline.} Our method significantly outperforms the baseline, even though the baseline uses the vertex order as additional information. Moreover, the baseline method does not generalize to point clouds with an arbitrary number of vertices.}
  \label{tab:table_MAIN}
  \vspace{-10pt}
\end{table*}

\paragraph{Datasets} 
We jointly train on the union of BEHAVE \cite{bhatnagar2022behave} and GRAB \cite{taheri2020grab}, obtaining a set of:
\begin{itemize}
    \item $15$ subjects, first $8$ subjects from the GRAB dataset and $7$ subjects from the official training part of BEHAVE;
    \item $40$ different classes of objects, including all $20$ objects from the BEHAVE dataset and $20$ selected objects from the GRAB dataset;
\end{itemize}
We downsample training sequences of GRAB and BEHAVE to $10fps$.
To evaluate our method, we select subjects $9$ and $10$ from the GRAB dataset and downsample the sequences to $30fps$. For the BEHAVE dataset, we use the official test part, which includes all sequences at $1fps$ with subject $3$ and part of the sequences with subjects $4,5$. 
As an input, we use point clouds with $9000$ points sampled uniformly over the SMPL-H \cite{pavlakos2019expressive} meshes. 
We refer to raw point clouds from the BEHAVE dataset used in our experiments as \emph{BEHAVE-Raw}. We use point clouds that are fused from 4 Kinect sensors and subsample $90k$ points from them.

\paragraph{Data augmentation.} 
During training, to simulate errors in the center prediction, we randomly translate and rotate the object around the ground-truth center $\objcentergt$.

\paragraph{Implementation details.} 
We implement our method using PyTorch framework and use Nvidia RTX3090 GPU for training and evaluation. 
The model is trained using Adam optimizer for $60$ epochs with a learning rate of $1e^{-4}$, which decays $10$ times after $30$-th and $40$-th epochs. 
For the first $20$ epochs of training, we use ground-truth object center $\objcentergt$ instead of predicted $\objcenter$ to select local neighborhood $\featpclocal$, to warm up the local PointNet++ encoder.

\begin{figure*}[!ht]
    \centering
    \begin{overpic}
    [trim=0cm 0cm 0cm 0cm,clip,width=\textwidth]{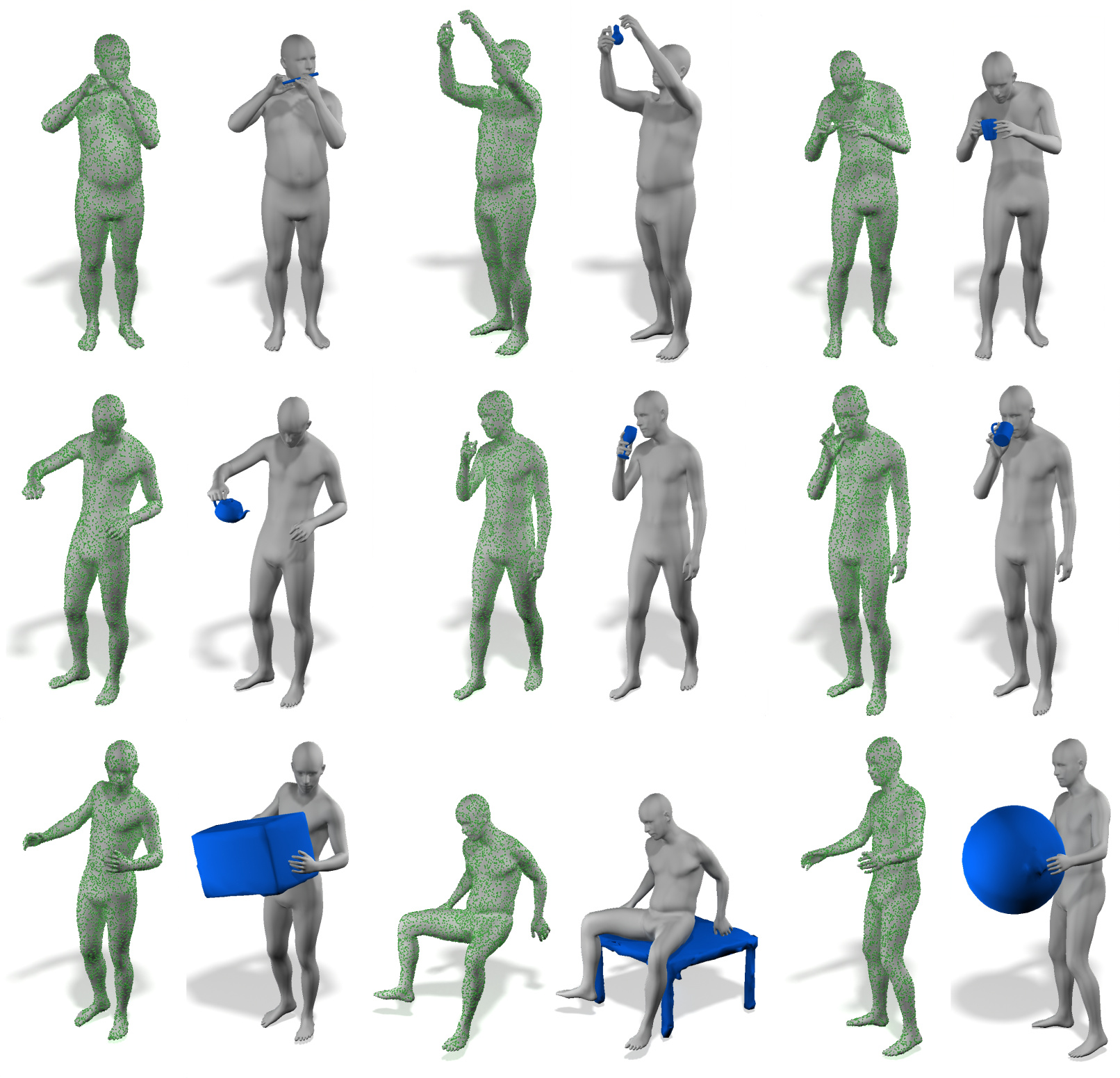}
    \end{overpic}
    \vspace{-0.8cm}
    \caption{\textbf{Qualitative results}. Results of our method on GRAB (first and second rows) and BEHAVE (third row) datasets.}
    \label{fig:virtual}
    \vspace{-15pt}
\end{figure*}

\paragraph{Nearest-Neighbor Baseline.} Since we are the first to tackle this task and no competitors are available, we propose a simple while informative baseline. 
Given the input point cloud, we recover the most similar in the training dataset in an L2 sense. 
Then, we recover the object handled by that subject and pose it in space in the same way. 
This baseline demonstrates that the task is non-trivial and the generalization to unseen poses and subjects of our method. 
Also, this baseline requires that the target point cloud and the ones in the training set share the same number of points. Hence, if the input point cloud is a raw scan, this baseline is not applicable. Our method, instead, does not rely on this assumption and is more general.

\paragraph{Object classification.} In our research, we also investigate the possibility of incorporating class prediction inside the network training. This task is significantly difficult at a single-frame level since an isolated pose often does not suggest a clear functionality. However, including this step is interesting to analyze the interaction and the nature of the network confusion. Hence, we modify our method by adding a decoder module that takes the global features $\featobjcenter$ and the local ones $\featpclocal$ as input to predict the object class. Then, we add to the training a simple cross-entropy loss between the predicted class and the ground truth one $\objclassgt$.

\subsection{Metrics}
In qualitative experiments, we use three main metrics to evaluate our results. 
In the tables, we report the average error across the considered test samples.

\paragraph{Vertex-to-vertex.} In most cases, our resulting object and the target one share the same number of vertices.
Hence, we can compute the error between our prediction and the ground truth $\widehat{\mathbf{T}}'$ as a point-to-point error:
\begin{equation}
    E_{v2v} = \| \ourout - \widehat{\mathbf{T}}'\|_F.
    \label{eq:v2v}
\end{equation}
When such error is computed only between the object centers, we will refer to it as $E_{c}$.

\paragraph{Chamfer distance.} When we evaluate the network that also predicts the class, target objects and selected templates might not share the same number of vertices. 
In that case, as a metric we use bi-directional Chamfer distance:
\begin{equation}
    E_{ch} = \frac{1}{\|\ourout\|} \sum_{x \in \ourout} \min_{y \in \widehat{\mathbf{T}}'} \|x - y \|_2 + \frac{1}{\|\widehat{\mathbf{T}}'\|} \sum_{y \in \widehat{\mathbf{T}}'} \min_{x \in \ourout} \|y - x \|_2
    \label{eq:chamfer}
\end{equation}

\paragraph{Classification Accuracy.} In case we use our network to predict the object class, we measure our misclassification error in terms of accuracy.

\subsection{Object pose Evaluation}
\begin{figure*}[ht!]
    \centering
    \begin{overpic}
    [trim=0cm 0.0cm 0cm 0cm,clip,width=\textwidth]{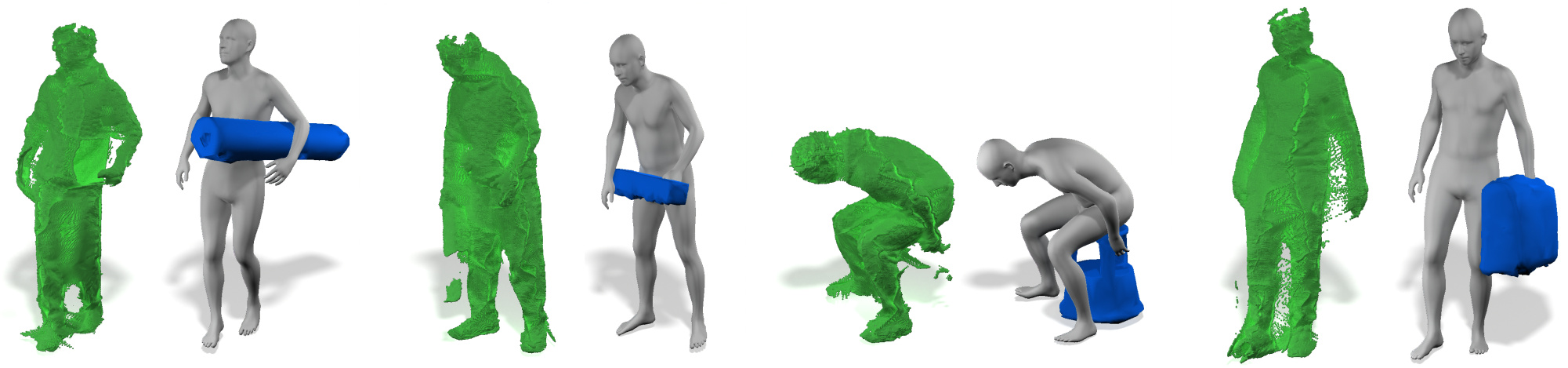}
    \end{overpic}
    \vspace{-0.5cm}
    \caption{\textbf{BEHAVE-Raw.} Results of our method considering as input raw point clouds from BEHAVE dataset. Despite the amount of noise and occlusions, our method is able to generalize and performs reliably.}
    \label{fig:RAW_PCS}
    \vspace{-10pt}
\end{figure*}

\begin{figure*}[ht!]
    \centering
    \begin{overpic}
    [trim=0cm 0cm 0cm 0cm,clip,width=\textwidth]{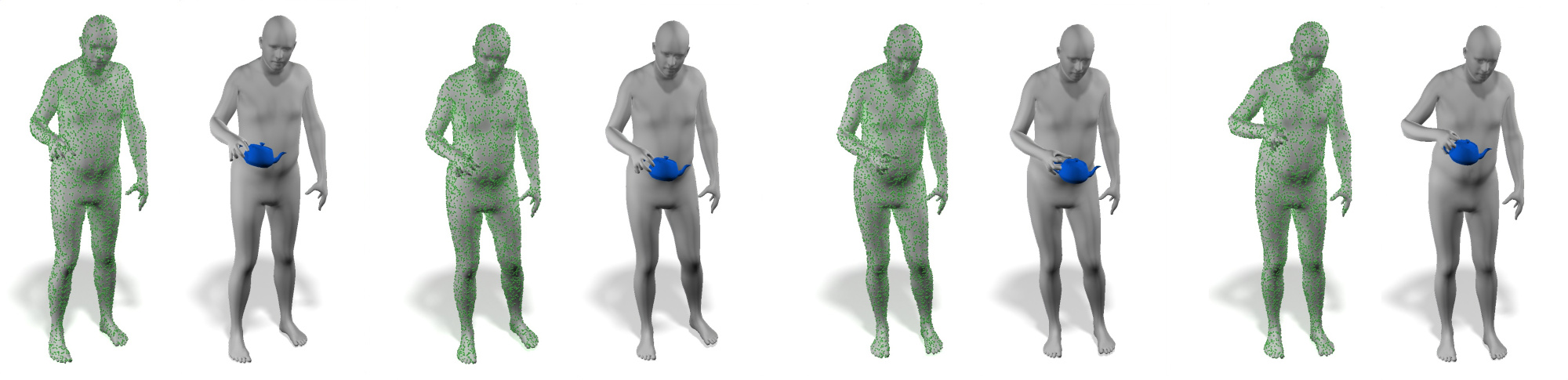}
    \end{overpic}
    \vspace{-0.7cm}
    \caption{\textbf{Human from IMUs.} Results of our method from a motion sequence acquired with IMUs. Even if the subject is unseen and the motion is subject to noise in the IMUs sensors, our method produces stable results.}
    \label{fig:IMUs}
    \vspace{-10pt}
\end{figure*}

In~\cref{tab:table_MAIN}, we report the quantitative evaluation on the test set of the datasets, comparing our method to the baseline. 
Our approach significantly outperforms the baseline, even if this latter exploits the points order information. 
We obtain the most significant margin on the GRAB dataset, where objects are small and mainly involve hands, showing the precision of our method. 
The baseline cannot be applied on BEHAVE-Raw since it does not share the same number of vertices as the training set, while our method shows only a limited performance decrease, pointing to generalization also to point clouds coming from different sources. 
We report qualitative results of our method on GRAB (first two rows) and BEHAVE (last row) in~\cref{fig:virtual}, and on BEHAVE-Raw in~\cref{fig:RAW_PCS}. 
Finally, to further evaluate the generalization of our method to unseen poses and subjects, we also considered point clouds obtained by an egocentric pipeline. 
In this scenario, we record a user motion with an XSens system \cite{roetenberg2007moven}, retarget the pose to an SMPL+H model, and obtain the point cloud by sampling the resulting mesh. Outputs of our method can be observed in~\cref{fig:results_imu} and ~\cref{fig:IMUs}. 
Generalizing to unseen poses and subjects acquired with wearable systems prone to measurement errors opens several exciting applications for VR/XR contexts.

\begin{table*}[ht!]
  \small
  \centering
  \begin{tabular}{lccccccccc}
    \multirow{2}{*}{\bf Methods} & \multicolumn{2}{c}{\bf GRAB} & \multicolumn{2}{c}{\bf BEHAVE} &  \multicolumn{2}{c}{\bf BEHAVE-Raw} \\
    \cline{2-7}
                                & $E_c\downarrow$  & $E_{v2v}\downarrow$ & $E_c\downarrow$ & $E_{v2v}\downarrow$ & $E_c\downarrow$ & $E_{v2v}\downarrow$ \\
    \hline
    Ours, hands\setfootnotemark\label{first}
                            & 0.0242 & \textbf{0.0871} & - & - & - & -  \\
    Ours, SMPL              & 0.0245 & 0.1009 & 0.0667 & \textbf{0.2877} & 0.0843 & 0.3205 \\
    Ours, SMPLH             & 0.0237  & 0.0943 & \textbf{0.0663} & 0.2900 & \textbf{0.0806}  & \textbf{0.3143} \\
    Ours, SMPLH+T           & \textbf{0.0235}  & 0.0983 & 0.0686 & 0.2880 & 0.0823 & 0.3161 \\
  \end{tabular}
  \caption{\textbf{Human Affordance.} We use our pipeline to explore how different inputs affect the object pose prediction. Using full body provides richer features for object recovery. }
  \label{tab:table_input}
  \vspace{-10pt}
  \afterpage{\afterpage{\footnotetext[\getrefnumber{first}]{The results were revised with the help of better data preprocessing.}}}
\end{table*}

\begin{table*}[ht]
  \small
  \centering
  \begin{tabular}{lcccccc}
    \multirow{2}{*}{\bf Method} & \multicolumn{3}{c}{\bf GRAB} & \multicolumn{3}{c}{\bf BEHAVE} \\
    \cline{2-7}
                                & $E_c\downarrow$ & $E_{ch}\downarrow$ & Acc.$\uparrow$ & $E_c\downarrow$ & $E_{ch}\downarrow$ & Acc.$\uparrow$ \\
    \hline
    NN, SMPLH               & 0.0404 & 0.1938 & 15.42 & 0.0880 & 0.3873 & 21.48 \\
    Ours, SMPLH             & 0.0290 & 0.1387 & 20.80 & 0.0728 & \textbf{0.3046} & 24.16 \\
    Ours, SMPLH + T         & \textbf{0.0263} & \textbf{0.1384} & \textbf{30.15}  & \textbf{0.0722} & 0.3069 & \textbf{49.28} \\
  \end{tabular}
  \vspace{-5pt}
  \caption{\textbf{Object classification}. Quantitative results of models with object class prediction. Introducing temporal information has a dramatic impact on object classification.}
  \label{tab:table_EXP_class}
  \vspace{-5pt}
\end{table*}

\begin{figure*}[!ht]
    \centering
    \begin{overpic}
    [trim=0cm 0.0cm 0cm 0cm,clip,width=0.95\linewidth]{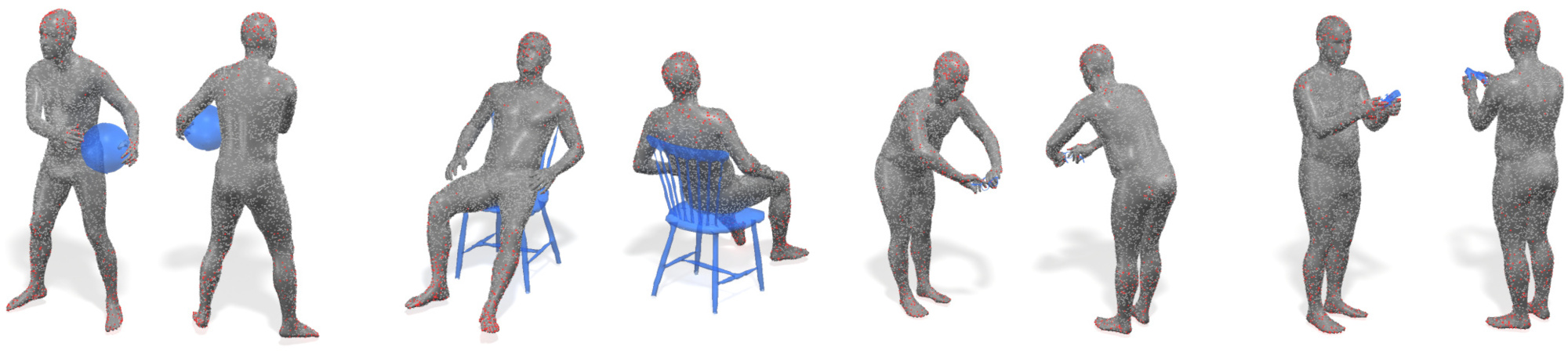}
    \scriptsize
    \put(7, -0.5){Basketball}
    \put(35, -0.5){Chairwood}
    \put(61, -0.5){Eyeglasses}
    \put(88, -0.5){Controller}
    \end{overpic}
    \caption{\textbf{Saliency.} Point cloud saliency computed for different objects, rendered from two perspectives. The contact region is relevant for all the interactions, while the network also focuses on the feet and head regions. All the predictions are results of our method.}
    \label{fig:saliency}
  \vspace{-10pt}
\end{figure*}

\begin{figure}[!ht]
    \centering
    \begin{overpic}
    [trim=0cm 0cm 0cm 0cm,clip,width=0.7\linewidth]{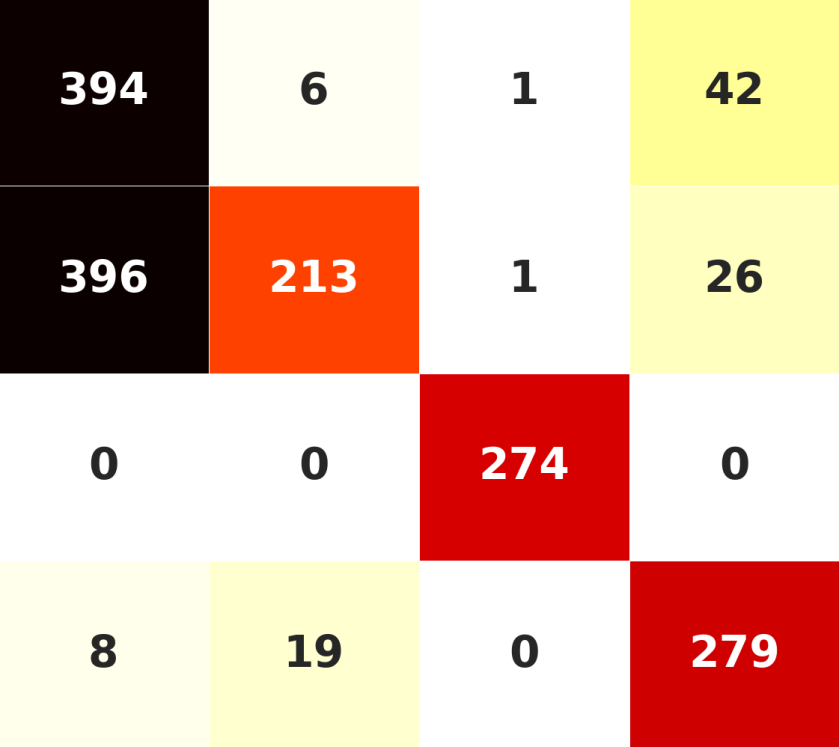}
    \scriptsize
    \put(4, 90){Coffeemug}
    \put(33, 90){Cup}
    \put(54, 90){Dorknoob}
    \put(81, 90){Hammer}
    
    \put(-4, 2){\rotatebox{90}{Hammer}}
    \put(-4, 25){\rotatebox{90}{Dorknoob}}
    \put(-4, 52){\rotatebox{90}{Cup}}
    \put(-4, 68.5){\rotatebox{90}{Coffeemug}}
    
    \end{overpic}
    \caption{\textbf{Confusion Matrix.} The confusion matrix for a subset of the predicted classes. Objects with similar functionality are often confused (Coffemug vs Cup), while ones that suggest characteristic human poses (Dorknoob, Hammer) are well separated.}
    \label{fig:confusion}
    \vspace{-18pt}
\end{figure}

\subsection{Human Affordance analysis}
We use our method to analyze human-object interaction considering three key factors: changes in the input information, the points saliency, and confusion in classification.

\paragraph{Input information.} Since we are curious to analyze how information is encoded in the inputs, we consider three more scenarios:
\begin{itemize}
    \item \textbf{Hands}: We generate the input data using the MANO \cite{romero2017mano} hand model annotations included in GRAB dataset. Hence, the model should infer the pose of the object without relying on other body parts.
    \item \textbf{SMPL}: We consider all the points of the subject, but the ones from the hands are in the rest pose provided by SMPL \cite{loper2015smpl}. In this way, we analyze how much the network captures the information of the finger pose.
    \item \textbf{SMPLH+T}: We consider all the points of the subject with the finger correctly posed (as in our method), and also we apply the temporal smoothing outlined in~\cref{sec:method}. Temporal information contextualizes the interactions and regularizes the predictions throughout the sequence. 
\end{itemize}
For the first two cases, we train ad-hoc networks. Results of this analysis are reported in~\cref{tab:table_input}.
We notice that hands provide crucial information for the GRAB dataset. The result is expected since, as the name suggested, all the interactions in the dataset are focused on hands. 
However, full-body context is crucial for reconstructing interactions from the BEHAVE dataset, as they often involve multiple body parts.

\paragraph{Points saliency.} As further evidence that object interaction involves different body parts, we conducted a study to discover what input points are crucial for the network. 
We follow a recent protocol to find 3D point cloud saliency \cite{zheng2019pointcloud}: we cast the input through the network, we compute a loss on the output (in our case, the one in ~\cref{eq:loss_off}), and we modify the input using the backpropagated gradient. 
This procedure is performed iteratively, and we refer to \cite{zheng2019pointcloud} for the details. 
We report the results of this study in~\cref{fig:saliency}, where points modified by the procedure are highlighted in red. 
We find the results of this analysis fascinating. 
As expected, the contact region is always essential to infer the correct object location. 
However, feet play a crucial role in all the reported cases, since they provide information about the human position and the consequent pose of the body. 
Another highlighted region is the head: different orientations give clues about object location and body posture.

\paragraph{Confusion in classification.} As a final analysis, we explore the learning object classification during the training. 
We jointly train a further MLP module that takes as an input $\featobjcenter$ and predicts the object class, using a cross-entropy loss. 
When a time sequence of point clouds is available, we exploit it by selecting the class with the highest score across the frames and applying it to the whole sequence. 
We report results in~\cref{tab:table_EXP_class}. 
Our experiment suggests that the task is challenging while, given the number of classes (40), we still consider our results a promising first step.
Also, we notice that temporal smoothing significantly helps classification accuracy. 
Temporal context disambiguate poses without a clear functionality. 
In~\cref{fig:confusion}, we report the confusion matrix for a subset of the classes for our method. 
The misclassification mainly arises from interactions of objects with similar functionality.

\section{Conclusions}
\label{sec:conclusions}
In this work, we have addressed a novel and inspiring problem that changes the perspective on object-human interaction. 
Our proposed model is simple, carefully designed, and inspired by behavioural studies. 
We collected evidence of the method's effectiveness on a large set of object classes and empirically proved its generalization on noisy and different inputs. 
Finally, our analysis of human affordance is unprecedented, showing that human-object interaction can also involve body parts distant from the object and pointing to interesting relations useful for applications and subsequent works.

\paragraph{Limitations and Future Works.} As the first exploration in this direction, our study enables several future possibilities. 
In this work, the temporal information is only used after the training procedure. 
Incorporating this information can create other patterns, further improving the results' quality. 
Our empirical evidence suggests that class prediction requires further investigation and more sophisticated techniques, like specialized attention mechanisms. 
Finally, we do not consider sequences that involve long (e.g., hours-long) and complex (e.g., multi-objects) interactions, which are difficult to capture. We hope our work can foster the community to collect such datasets.

{\footnotesize
\paragraph{Acknowledgements}
Special thanks to the RVH team and reviewers, their feedback helped improve the manuscript. We also thank Omid Taheri for the help with GRAB objects.
This work is funded by the Deutsche Forschungsgemeinschaft - 409792180 (EmmyNoether Programme, project: Real Virtual Humans) and the German Federal Ministry of Education and Research (BMBF): Tübingen AI Center, FKZ: 01IS18039A. 
G. Pons-Moll is a member of the Machine Learning Cluster of Excellence, EXC number 2064/1 – Project number 390727645. 
The authors thank the International Max Planck Research School for Intelligent Systems (IMPRS-IS) for supporting I. Petrov. R. Marin is supported by an Alexander von Humboldt Foundation Research Fellowship. The project was made possible by funding from the Carl Zeiss Foundation.
}

{\small
\bibliographystyle{ieee_fullname}
\bibliography{egbib}
}

\newpage
{\noindent \large \bf {APPENDIX}}\\

This appendix, provides an ablation study on our design choices in \cref{sec:sup_ablation}. In \cref{sec:sup_architecture} we report further details on our method, how we modified it to address the class prediction task, and a description of the considered baseline. In \cref{sec:sup_results} we include results for different kinds of inputs (shapes from GRAB and BEHAVE, IMUs, and noisy point clouds), and discuss failure cases oth the method. Finally, in \cref{sec:sup_affordance} we describe the points saliency estimation procedure, report further examples, and include the full confusion matrix for the classification.

\setcounter{table}{0}
\setcounter{figure}{0}
\renewcommand{\thefigure}{A.\arabic{figure}}
\renewcommand{\thetable}{A.\arabic{table}}
\appendix
\section{Ablation Study}
\label{sec:sup_ablation}
\begin{table}[h!]
  \small
  \centering
  \begin{tabular}{lcccc}
    \multirow{2}{*}{\bf Method} & \multicolumn{2}{c}{\bf GRAB} & \multicolumn{2}{c}{\bf BEHAVE} \\
    \cline{2-5}
                                & \bf{$E_c$}  & \bf{$E_{v2v}$}& \bf{$E_c$}  & \bf{$E_{v2v}$} \\
    \hline
    R,t                 & 0.0727 & 0.2556 & 0.0804 & 0.3167 \\
    no $\featpclocal$   & 0.0260 & 0.1112 & \textbf{0.0617} & \textbf{0.2868} \\
    \hline
    Ours                & \textbf{0.0237} & \textbf{0.0943} & 0.0663 & 0.2900 \\  
     \\
  \end{tabular}
  \caption{\textbf{Ablation study.} We ablate different parts of our method, showing their importance for the quality of the final results. We observe that directly predicting the rotation and the translation produces the worst results.}
  \label{tab:supp_table_ABL}
  \vspace{-10pt}
\end{table}

We perform an ablation study to validate our design choices and analyze their impact on the system. We report the results in~\cref{tab:supp_table_ABL}.

\vspace{0.2cm}\textbf{R,t.} Our method outputs a vertex-wise offset. Given that our goal is recovering a global pose of a rigid object, a more natural option is to predict a global rotation and translation directly. However, training a model this way significantly decreases performance, suggesting that our richer output representation provides the network more flexibility.

\vspace{0.2cm}\textbf{No $\featpclocal$.} One of our hypotheses is that the correct object location arises from the union of human parts. In our design, we enrich the features from the whole body with others that are focused around the predicted center location of the object. By removing the latter, we observe up to $15\%$ impact on the performance. We discover this information is particularly relevant for small objects (i.e., GRAB\cite{taheri2020grab}), where recovering the pose requires a finer understanding. The performance for the larger objects (i.e., BEHAVE\cite{bhatnagar2022behave}) is on par with the main model, as in this case, the whole body becomes predominant in the prediction.

\section{Architecture and implementation details}
\label{sec:sup_architecture}
In this section we provide more details about the proposed architecture. We gather the used notations in \cref{tab:supp_table_NOTATION}.

\begin{table}[ht!]
    \begin{center}
    \begin{tabular}{l|l}
    \small
    \textbf{Symbol}                                         & \textbf{Meaning}                            \\
    $\pc            \in \mathbb{R}^{n \times 3}$            & Input 3D human point cloud                  \\   
    $\objcenter     \in \mathbb{R}^3$                       & Predicted Object Center                     \\                   
    $\objcentergt   \in \mathbb{R}^3$                       & Ground truth Object Center                  \\                 
    $\pclocal       \in \mathbb{R}^{2000 \times 3}$         & Selected neighbourhood                      \\                   
    $\featobjcenter \in \mathbb{R}^{512}$                   & Features from the global point cloud        \\         
    $\featpclocal   \in \mathbb{R}^{128}$                   & Features from the local neighbourhood       \\         
    $\classes       = \{c_1, c_2, \dots, c_{39} \}$         & Classes                                     \\        
    $\objclass \in \classes$                                & Object class (given as input, or predicted) \\
    $\objkeyp \in \mathbb{R}^{1500 \times 3}$               & Key points of the class  $\objclass$        \\    
    $\objtemplate \in \mathbb{R}^{m \times 3}$              & Template for the class  $\objclass$         \\
    $\offset \in \mathbb{R}^{1500 \times 3}$                & Predicted  point-wise offset                \\
    $\offsetgt \in \mathbb{R}^{1500 \times 3}$              & Ground truth  point-wise offset             \\
    
    \end{tabular}
    \end{center}
    \caption{\textbf{Symbols.} In this table we report the main symbols used across the method.}
\label{tab:supp_table_NOTATION}
\vspace{-10pt}
\end{table}

\subsection{Object pop-up}

\paragraph{Training details.}
Training the model for $60$ epochs on Nvidia RTX3090 GPU takes approximately  $18$ hours. Sequences from the GRAB dataset are downsampled from \emph{120fps} to \emph{10fps} for training and \emph{30fps} for evaluation. The BEHAVE dataset provides extended annotations at \emph{30fps} for some sequences, that are downsampled to \emph{10fps} for training. For evaluation on BEHAVE, the original \emph{1fps} annotations are used. We align the meshes to share the same ground plane, keeping the original global rotation.

\begin{figure*}[!ht]
\centering
    \begin{overpic}[trim=0cm 0cm 0cm 0cm,clip,width=1.0\linewidth]{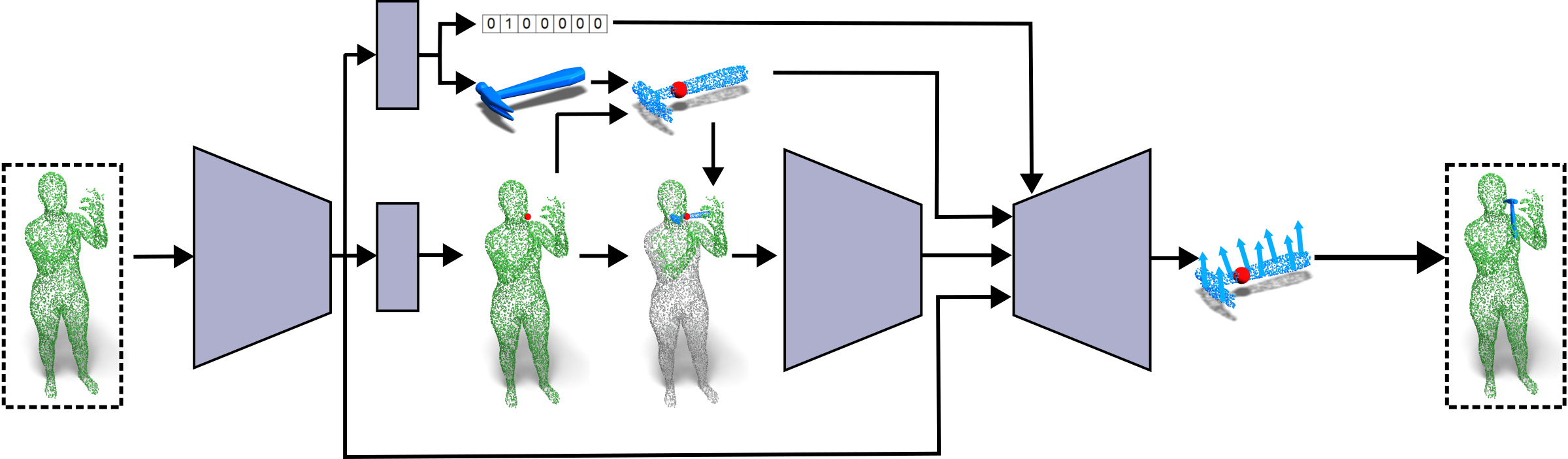}
    \scriptsize
    
    \put(0.5,2){Point Cloud}
    \put(3.5,0.7){$\mathbf{P}$}    

    \put(30,25.5){Template $\objtemplate$}    
    \put(40,25.5){Key points $\objkeyp$}       
    
    \put(30,2.2){Object Center}
    \put(33,0.7){$\objcenter$}        

    \put(41,2.2){Local Neigh.}
    \put(44,0.7){$\pclocal$}  
    
    \put(77,6){Key points}
    \put(78.3,4){offsets}    
    \put(78.8,2){$\offset$}  
    
    \put(85,16){Template}  
    \put(85.5,14){Fitting}  
    
    \put(94,1){Output}  
    
    \put(13.5,12){PointNet++}  
    \put(51.5,12){PointNet++}  
    \put(66.5,12){Decoder} 
    
    \put(60.5,16){$\objkeyp$}
    \put(60.5,8.5){$\featobjcenter$} 
    \put(59.8,13.7){$\featpclocal$} 
    \put(66.2,19){$\objclass$} 
    
    \put(24.7,14.4){\rotatebox{-90}{MLP}}
    \put(24.7,27.3){\rotatebox{-90}{MLP}}
    \end{overpic}
\caption{\label{fig:pipeline_class} \textbf{Object pop-up with class prediction.} Our modified method predicts the object position and the class, starting only from an input point cloud. We train another module to take as input $\featobjcenter$ and predict the class $c$ for the considered interaction.}
\end{figure*}
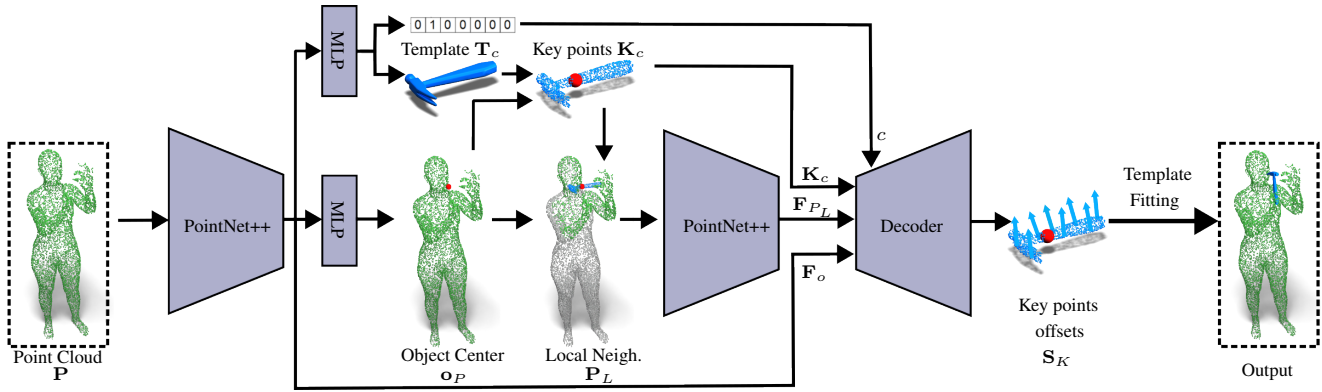

\subsection{Object pop-up with class prediction}
The overview of the model with class prediction is presented in \cref{fig:pipeline_class}. Object class is predicted from global $\featobjcenter$ using MLP, apart from that module other modules are similar to the main model. 

\paragraph{Training details.}
This model is trained in exactly the same setting as the model without class prediction. We use cross entropy loss $L_{cel}$, to supervise class prediction.

The network is then trained using the following loss:
\begin{equation}
    L = L_{\objcenter} + \alpha L_{off} + L_{cel}.
\end{equation}
The weighting coefficient is $\alpha = 100$.

\subsection{Nearest neighbour baseline}
Here we provide a detailed description of the Nearest Neighbour baseline. First of all, we consider the set of training point clouds 
\begin{equation}
    \mathbf{P}^{train} = \{\mathbf{P}^{train}_1, \mathbf{P}^{train}_2, \dots, \mathbf{P}^{train}_k\},
\end{equation}
where each $\mathbf{P}^{train}_i$ is equipped with a correctly posed object template $\mathbf{T}^{train}_i$ and the relative class $c^{train}_i$. 
Then, given a new input point cloud $\pc$, we look for the closest point cloud in the training set:

\begin{equation}
    \hat{i} = \argminB_{i \in [0, \dots, k]} \|\mathbf{P} - \mathbf{P}^{train}_i \|_F.
\end{equation}
When the task is to recover the object with the class given as input, we consider only the training samples of the given class, and we retrieve the object $\mathbf{T}^{train}_{\hat{i}}$. When the method also has to predict the class, we consider the entire training dataset, and we also output the associated class $c^{train}_{\hat{i}}$.

\vspace{0.2cm} \textbf{Remark.} This baseline can be applied only if the input point cloud shares the same number and order of vertices as the ones coming from the training set. While this gives an advantage to the baseline, we decided to proceed this way for a computational reason, given the large training dataset size.

\section{Results}
\label{sec:sup_results}
This section presents more qualitative results of the proposed Object pop-up method and discusses the failure cases.

\begin{figure*}[!ht]
    \centering
    \includegraphics[trim=0cm 0cm 0cm 0cm,clip,width=\linewidth]{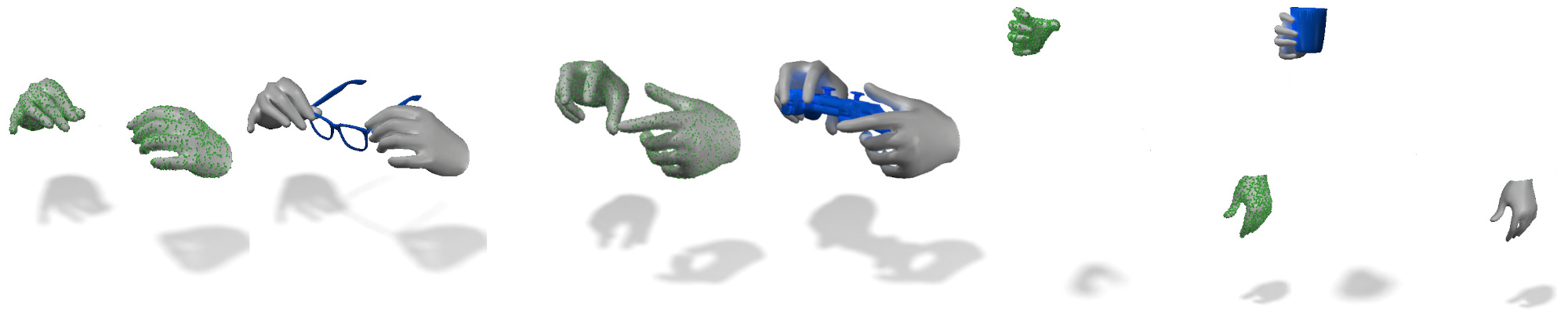}
    \caption{\textbf{Hands.} Our method results on hands from the GRAB dataset. The input point cloud is visualized over the human mesh for reference. The prediction is made using just a point cloud and an object class label.}
    \label{fig:hands_sm}
    \vspace{10pt}
\end{figure*}

\begin{figure*}[!h]
    \centering
    \includegraphics[trim=0cm 0cm 0cm 0cm,clip,width=\linewidth]{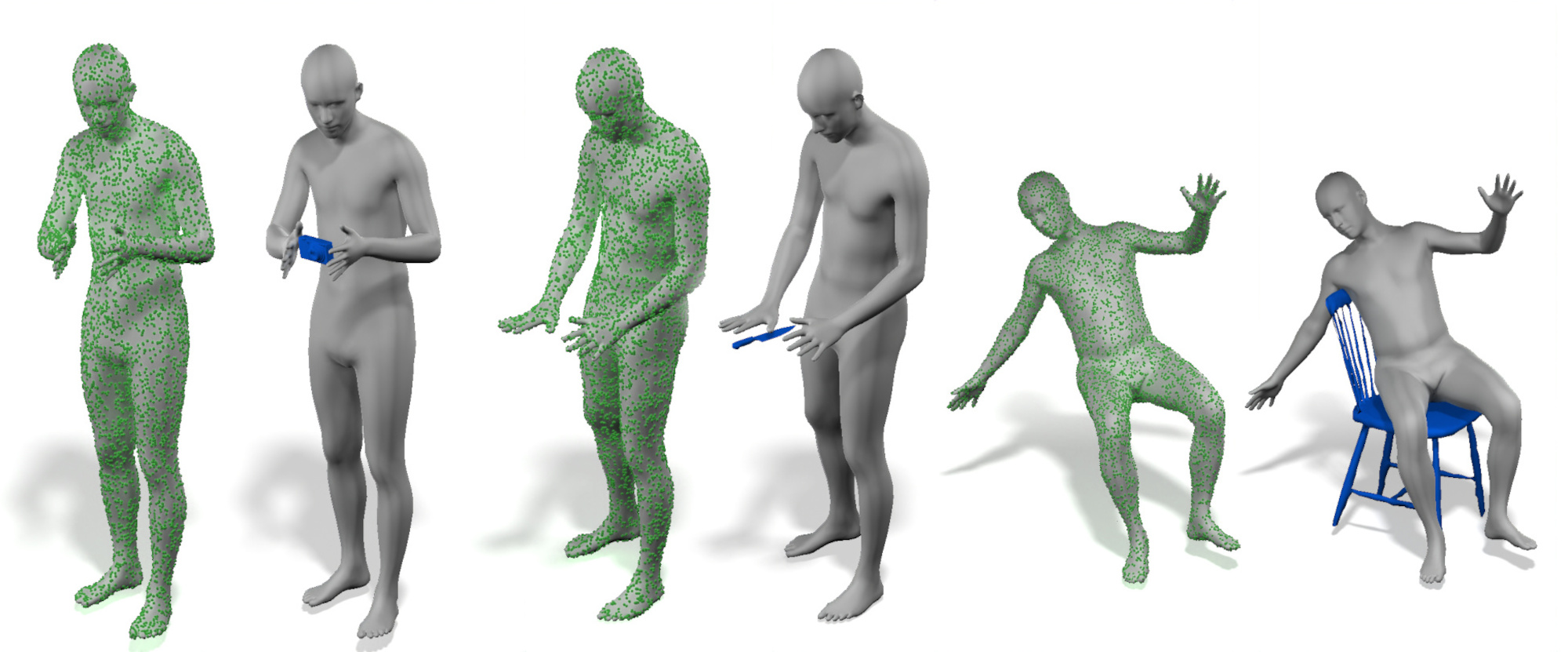}
    \caption{\textbf{SMPL.} Results of our method on data generated from SMPL model from GRAB and BEHAVE. This case differs from data sampling from SMPL-H model by the absence of articulated hand pose. The input point cloud is visualized over the human mesh for reference. The prediction is made using just a point cloud and an object class label.}
    \label{fig:smpl_sm}
    \vspace{10pt}
\end{figure*}

\paragraph{Hands and SMPL.}
Qualitative results for Object pop-up trained on data generated from MANO \cite{romero2017mano} hand meshes from the GRAB dataset are presented in \cref{fig:hands_sm}. 
The vast majority of actions in GRAB are done with hands, so that the model can predict the object's position quite well. 
However, for other types of interactions (i.e. involving other body parts), having full-body context is crucial. 
In \cref{fig:smpl_sm} we show qualitative results for Object pop-up model trained on data generated from SMPL \cite{loper2015smpl} meshes from GRAB and BEHAVE. This data differs from the primary training data generated from SMPL-H \cite{pavlakos2019expressive} meshes by the absence of articulated hand pose. Some easier interactions with objects through hands, like lifting a camera on the left side of \cref{fig:smpl_sm}, are handled by the model well. However, more complex cases, like cutting with a knife in the middle of \cref{fig:smpl_sm}, lead to erroneous prediction because the model lacks local features essential for the interaction. At the same time, interactions involving full-body, like sitting in the right of \cref{fig:smpl_sm}, are perceived well by the model because hands are not contributing much to the local interaction context there.

\begin{figure*}[!ht]
    \vspace{20pt}
    \centering
    \includegraphics[trim=0cm 0cm 0cm 0cm,clip,width=\linewidth]{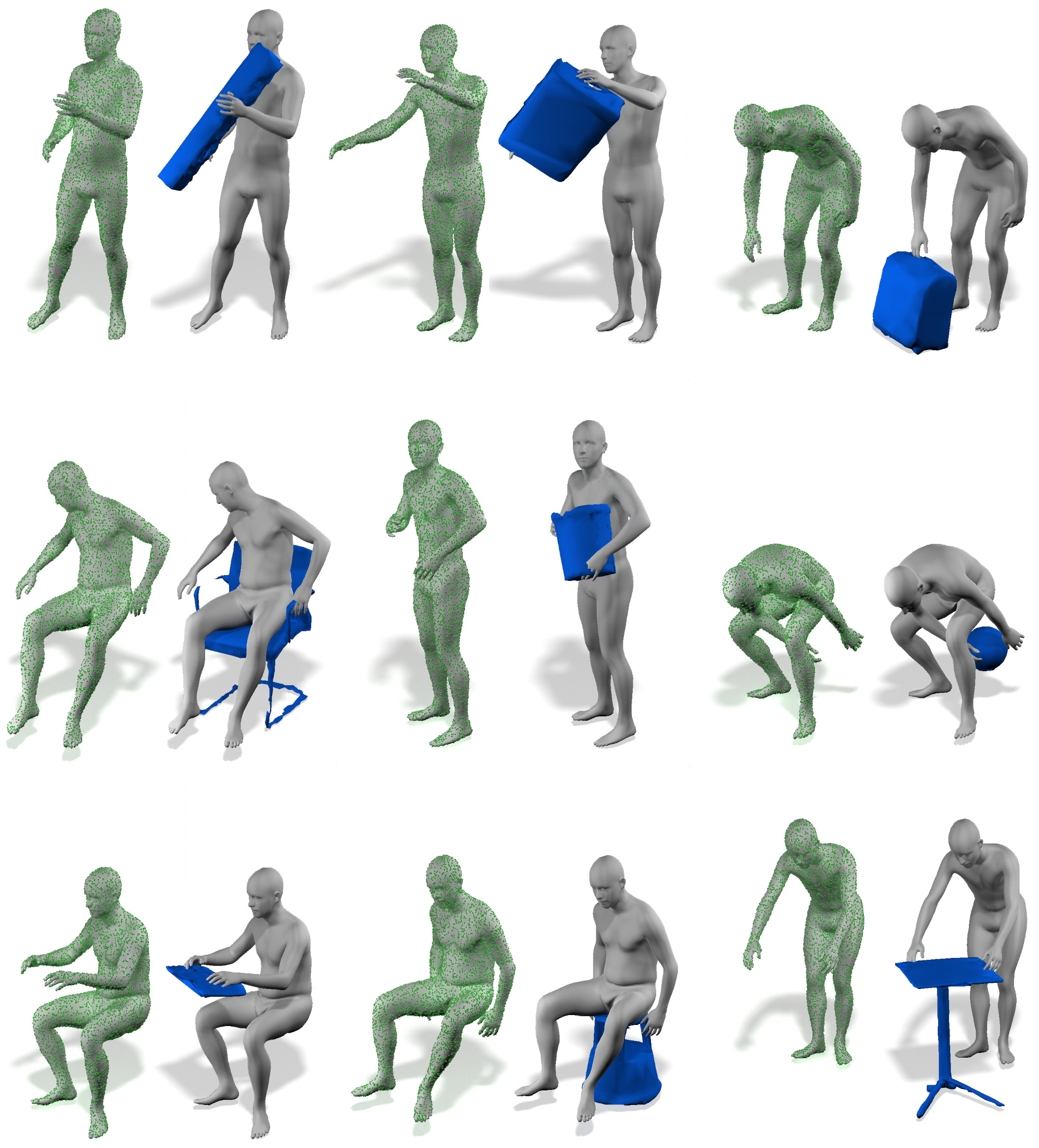}
    \caption{\textbf{BEHAVE.} Results of our method on the BEHAVE dataset. The input point cloud is visualized over the human mesh for reference. The prediction is made using just a point cloud and an object class label.}
    \label{fig:behave_sm}
    \vspace{20pt}
\end{figure*}

\begin{figure*}[!ht]
    \vspace{20pt}
    \centering
    \includegraphics[trim=0cm 0cm 0cm 0cm,clip,width=\linewidth]{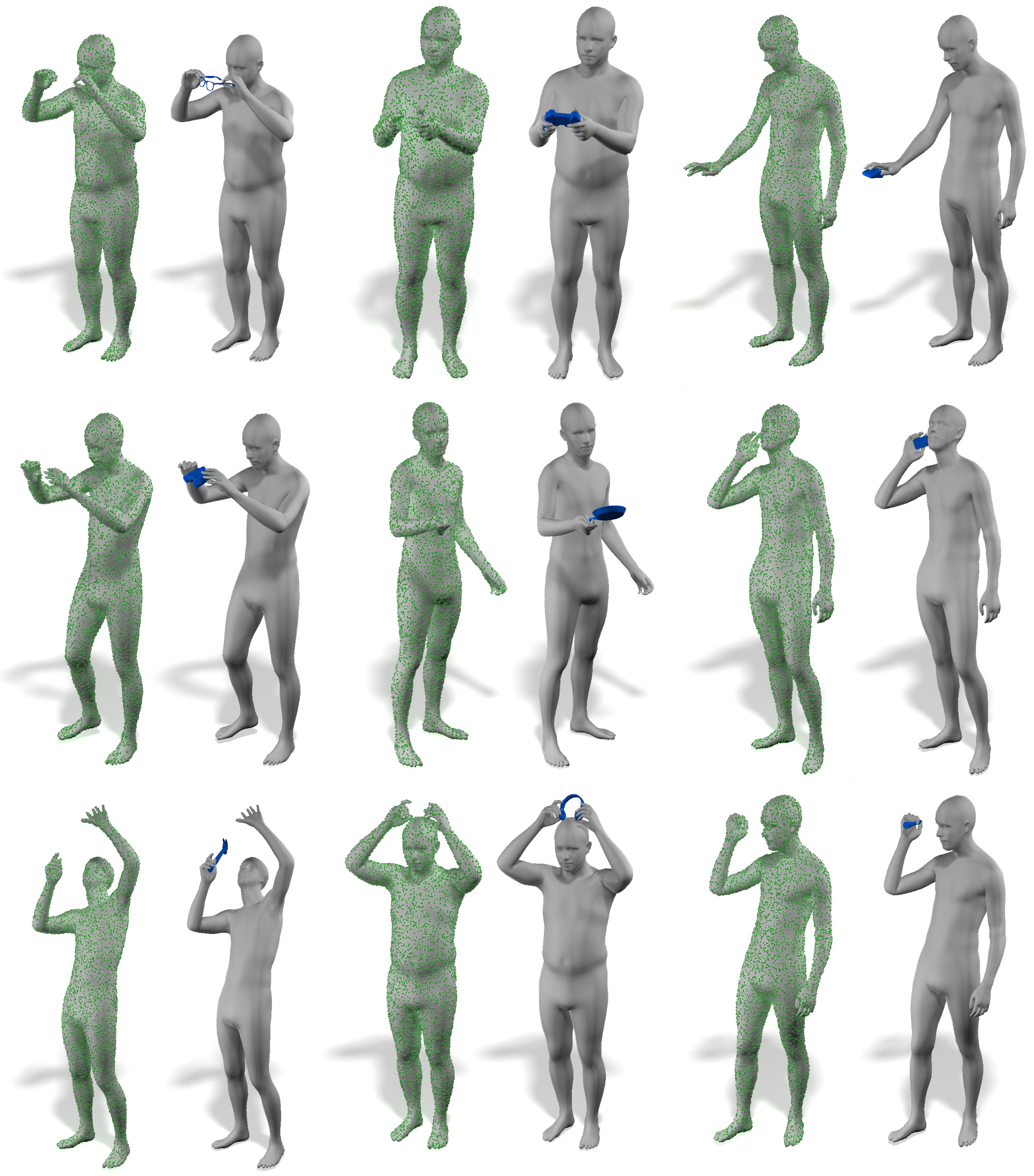}
    \caption{\textbf{GRAB.} Results of our method on the GRAB dataset. The input point cloud is visualized over the human mesh for reference. The prediction is made using just a point cloud and an object class label.}
    \label{fig:grab_sm}
    \vspace{20pt}
\end{figure*}

\paragraph{GRAB and BEHAVE.}
We present more qualitative results on GRAB in \cref{fig:grab_sm} and BEVAHE in \cref{fig:behave_sm}. 
Our method can predict the object's realistic location for many classes.

\begin{figure*}[!h]
    \centering
    \includegraphics[trim=0cm 0cm 0cm 0cm,clip,width=\linewidth]{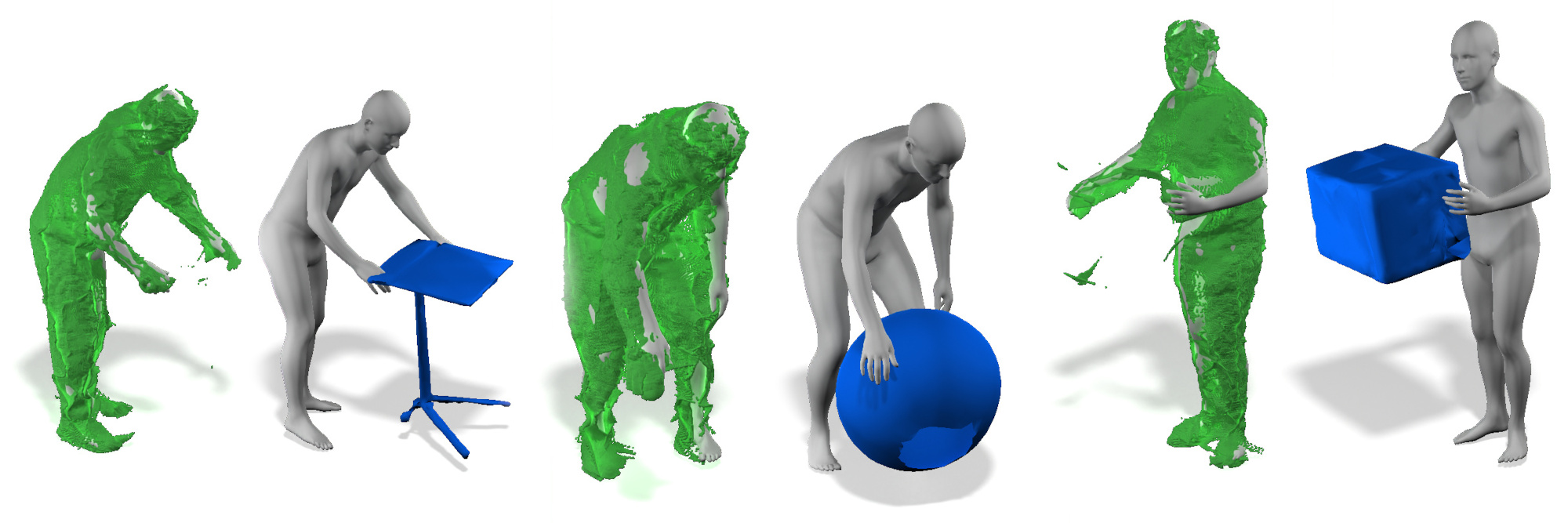}
    \caption{\textbf{BEHAVE-Raw.} Results of our method on raw point clouds from the BEHAVE dataset. The input point cloud is visualized over the human mesh for reference. The prediction is made using just a point cloud and an object class label.}
    \label{fig:rawpc_sm}
    \vspace{10pt}
\end{figure*}

\begin{figure*}[!ht]
    \centering
    \begin{overpic}
    [trim=0cm 0cm 0cm 0cm,clip,width=\linewidth]{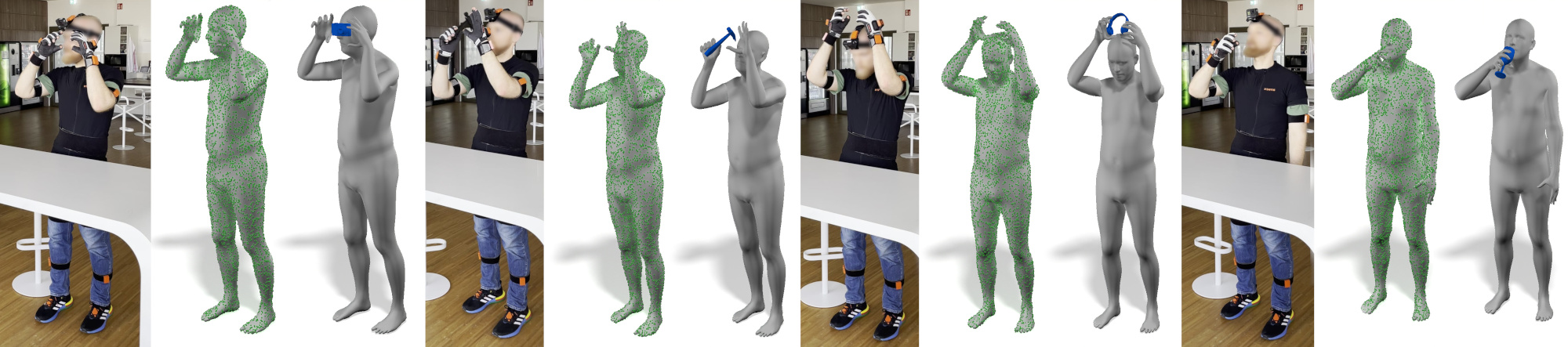}
    \scriptsize
    \put(1.7,24){Ref. Image}
    \put(2.0,22.5){(not used)}
    \put(13,24){Input}
    \put(11,22.5){Point Cloud}
    \put(20.5,24){Result}

    \put(27.5,24){Ref. Image}
    \put(27.8,22.5){(not used)}
    \put(38,24){Input}
    \put(36,22.5){Point Cloud}
    \put(45.5,24){Result}

    \put(51.7,24){Ref. Image}
    \put(52,22.5){(not used)}
    \put(62,24){Input}
    \put(60,22.5){Point Cloud}
    \put(69.3,24){Result}

    \put(76.3,24){Ref. Image}
    \put(76.6,22.5){(not used)}
    \put(87,24){Input}
    \put(85,22.5){Point Cloud}
    \put(94.5,24){Result} 

    \end{overpic}
    \caption{\textbf{Human from IMUs.} Results of our method on data acquired with wearable IMUs. The input point cloud is visualized over the human mesh for reference. The prediction is made using just a point cloud and an object class label.}
    \label{fig:imus_sm}
\end{figure*}

\paragraph{Generalization.}
We present additional results on BEHAVE-Raw in \cref{fig:rawpc_sm}, and in \cref{fig:imus_sm} we report more qualitative examples on the data recorded with IMU sensors. 
Our method shows generalization to both noisy point clouds of BEHAVE-Raw and unseen subjects, recorded with a wearable IMU setup. 

\paragraph{Failure cases.}

\begin{figure*}[!ht]
    \centering
    \includegraphics[trim=0cm 0cm 0cm 0cm,clip,width=\linewidth]{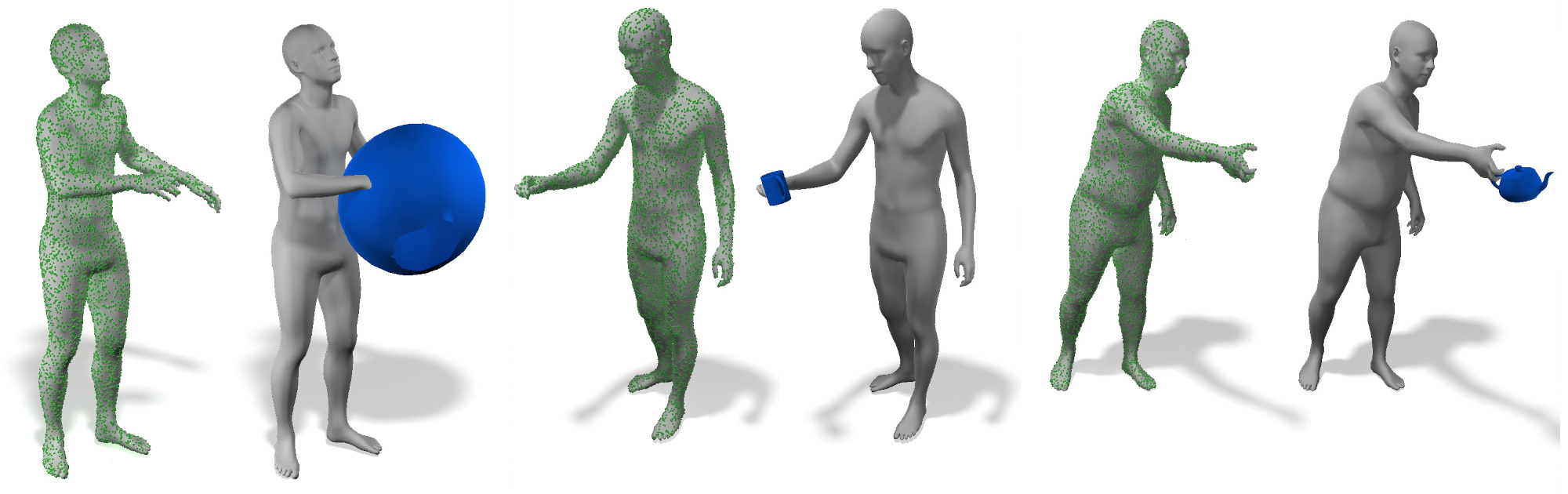}
    \caption{\textbf{Failure cases.} Our method may fail to predict correct object location. (Left) Interpenetration of the object and human might happen, because the method does not fully account for a surface of the human body. (Mid) Incorrect position for an object that can be grasped both by the handle and by the body with only slight variation in hand pose. (Right) Wrong object location for non-object-specific interactions, such as pass, lift or inspect.}
    \label{fig:failure_sm}
\end{figure*}

Our method sometimes predicts objects interpenetrating the human body (e.g. yoga ball on the left of \cref{fig:failure_sm}). The absence of an explicit surface in the input data requires the network to recover such a complex structure and the non-interpenetrating relationship.
In some cases, the method struggles to predict correct object placement for objects with small handles (e.g. mug in the middle of \cref{fig:failure_sm}, as grabbings by different parts (e.g., by the handle, or by the central body of the object) are all plausible solutions. 
These two cases differ only in a slight variation of hand pose, which can be hard to grasp even with the local focus of the network. 
Another failure case is incorrect object placement for interactions that do not involve objects' functionality (i.e. lifting, passing, inspecting). 
An example of such a case is a teapot interaction on the right side of \cref{fig:failure_sm}, where it is just shifted in the space. 
For this case, the method still predicts an object pose which is more common for the object's functionality (\cref{fig:grab_sm} presents such distinctive teapot grasp).

\section{Human affordance details}
\label{sec:sup_affordance}
\subsection{Points saliency estimation}
Here we report the details about Points saliency estimation, following the procedure from the Algorithm $1$ of \cite{zheng2019pointcloud}. 

\begin{figure*}[!ht]
    \centering
    \begin{overpic}
    [trim=0cm 0cm 0cm 0cm,clip,width=\linewidth]{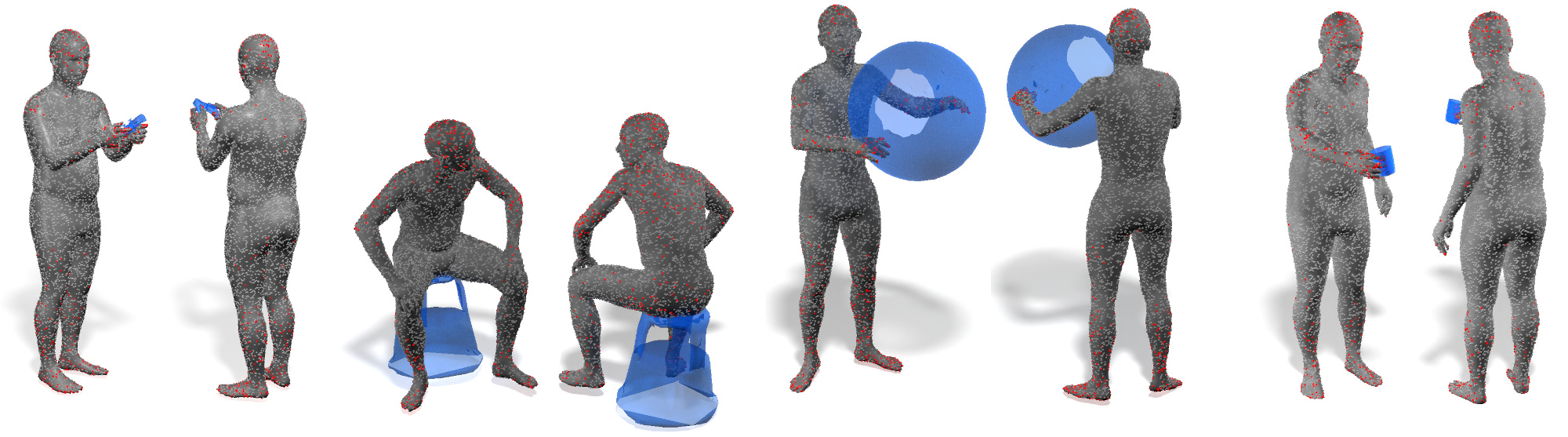}
    \scriptsize
    \put(6, -1){Game controller}
    \put(33, -1){Stool}
    \put(60, -1){Yoga ball}
    \put(89.5, -1){Cup}
    \end{overpic}
    \caption{\textbf{Saliency.} Point cloud saliency computed for different object predictions, rendered from two different perspectives. We observe that contact region is relevant for all the interaction, while the network also focuses on feet and head. All the objects are results of our method.}
    \label{fig:saliency_sm}
  \vspace{-5pt}
\end{figure*}

\begin{enumerate}
    \item Given an input point cloud $\pc$ and the associated class $\objclass$, first of all we compute the center of $\pc$ as the median of the three individual coordinates of the points: 
    \begin{equation}
        \mathbf{p}_m = (median(\mathbf{P}_x), median(\mathbf{P}_y), median(\mathbf{P}_z)).
    \end{equation}
    \item For each point, we compute the vector that connect the center $\mathbf{p}_m$ to it:
    \begin{equation}
        \mathbf{r}_i = (\mathbf{p}_i - \mathbf{p}_m)
    \end{equation}
    \item We cast $\pc$ and $\objclass$ through the network, obtaining the output offsets $\offset$. We use them to compute $L_{off}$ as reported in the main manuscript.
    \item For each input point $\mathbf{p}_i$ of the point cloud, we recover the gradient by backpropagation:
    \begin{equation}
        \mathbf{g}_i = \nabla_{p_i} L_{off}
    \end{equation}
    \item We construct the point-wise saliency map as:
    \begin{equation}
        s_i = - \|\mathbf{r}_i\|_2 (\mathbf{r}_i \cdot \mathbf{g}_i)
    \end{equation} 
    \item We pick the $90$ input points ($1\%$ of the point cloud) associated to the top saliency scores, and  we shift the position of each of these points toward the shape median:
    \begin{equation}
        \tilde{\textbf{p}}_j = \textbf{p}_j - 0.05 \mathbf{r}_i  
    \end{equation}
    \item We substitute these values in the original point cloud, and we restart the procedure from the beginning for $10$ times.
\end{enumerate}
In~\cref{fig:saliency_sm} we report further results of this procedure (in red, the points touched by these iterations).

\subsection{Confusion matrix of classes.}
\begin{figure*}[!ht]
    \centering
    \begin{overpic}
    [trim=0cm 0cm 0cm 0cm,clip,width=\linewidth]{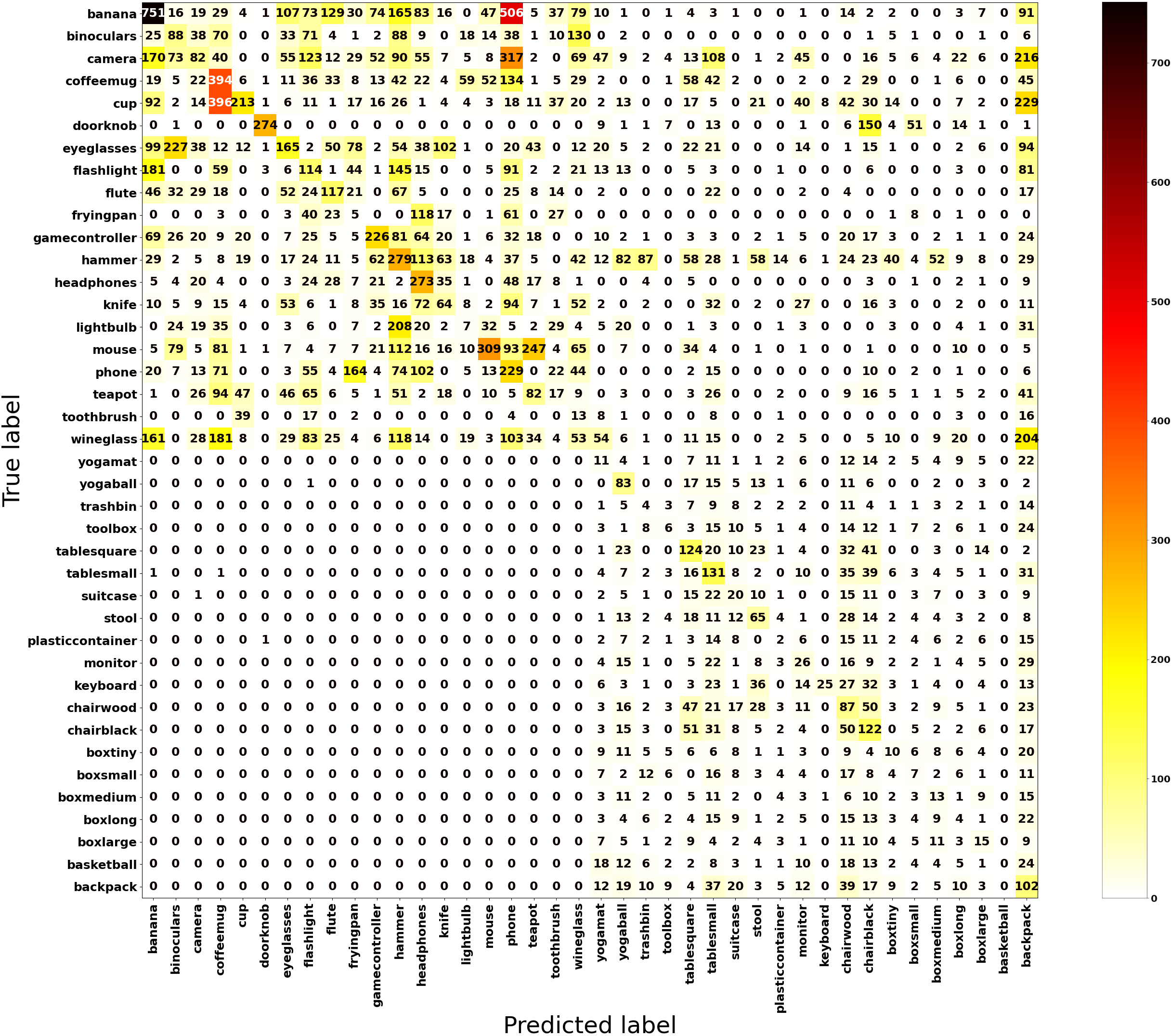}
    \scriptsize
    \end{overpic}
    \caption{\textbf{Confusion Matrix.} Here we report the full confusion matrix for the classification prediction. Zoomin for details.}
    \label{fig:confusion_sm}
  \vspace{-10pt}
\end{figure*}

For the sake of completeness, in~\cref{fig:confusion_sm} we report the full confusion matrix for all the classes on the classification task. We observe that the task is particularly challenging, and several ambiguous cases exist. We believe this is due to the presence of objects with similar functions (e.g. various boxes, two chairs and a stool, etc.), but also the ambiguity of some datasets sequence (e.g., a human inspecting object without actually using it with a clear functionality). In future, the collection of other datasets designed explicitly for this task will significantly ease the learning.

\end{document}